\documentclass{article}
\pdfoutput=1

\usepackage[preprint]{neurips_2019}

\usepackage[utf8]{inputenc} 
\usepackage[T1]{fontenc}    
\usepackage{hyperref}       
\usepackage{url}            
\usepackage{booktabs}       
\usepackage{amsfonts}       
\usepackage{nicefrac}       
\usepackage{microtype}      
\usepackage[ruled]{algorithm2e}
\usepackage{tikz}

\usepackage{caption}
\usepackage{subcaption}
\usepackage{lmodern}

\title{ChemGrapher: Optical Graph Recognition of Chemical Compounds by Deep Learning}

\author{Martijn ~Oldenhof\\
  ESAT - STADIUS\\
  KU Leuven\\
  Leuven, 3001, Belgium \\
  \texttt{martijn.oldenhof@esat.kuleuven.be} \\
  \And
  Adam ~Arany\\
  ESAT - STADIUS\\
  KU Leuven\\
  Leuven, 3001, Belgium \\
  \texttt{adam.arany@esat.kuleuven.be} \\
  \And
  Yves ~Moreau\\
  ESAT - STADIUS\\
  KU Leuven\\
  Leuven, 3001, Belgium \\
  \texttt{moreau@esat.kuleuven.be} \\
  \And
  Jaak ~Simm\\
  ESAT - STADIUS\\
  KU Leuven\\
  Leuven, 3001, Belgium \\
  \texttt{jaak.simm@esat.kuleuven.be}}

\begin{document}
\setcitestyle{numbers}
\newtheorem{definition}{Definition}
\newtheorem{assumption}{Assumption}
\setcitestyle{square}
\maketitle

\begin{abstract}
In drug discovery, knowledge of the graph structure of chemical compounds is essential. Many thousands of scientific articles in chemistry and pharmaceutical sciences have investigated chemical compounds, but in cases the details of the structure of these chemical compounds is published only as an images. A tool to analyze these images automatically and convert them into a chemical graph structure would be useful for many applications, such drug discovery. A few such tools are available and they are mostly derived from optical character recognition. However, our evaluation of the performance of those tools reveals that they make often mistakes in detecting the correct bond multiplicity and stereochemical information. In addition, errors sometimes even lead to missing atoms in the resulting graph.  In our work, we address these issues by developing a compound recognition method based on machine learning. More specifically, we develop a deep neural network model for optical compound recognition. The deep learning solution presented here consists of a segmentation model, followed by three classification models that predict atom locations, bonds and charges. Furthermore, this model not only predicts the graph structure of the molecule but also produces all information necessary to relate each component of the resulting graph to the source image. This solution is scalable and could rapidly process thousands of images. Finally, we compare empirically the proposed method to the well-established tool OSRA~\cite{osra} and observe significant error reductions.
\end{abstract}
\section{Introduction}

Knowledge of the chemical structure of compounds is central in drug discovery because this structure determines the properties of the compound. It is for example used for drug candidate selection. Because billions of euros of research and development investment are needed to successfully bring a new drug to the market, any tool that improves the drug candidate selection process would have a significant pharmaceutical impact. 

Although chemical structures, which are the familiar graph drawings of molecules, do lose some information about the electronic structure of a molecule (which is actually responsible for its chemical properties), they are powerful and effective abstractions. To query such structures or apply machine learning, we need to start from a well-structured data set encoding the graph representation of the chemical structure. This encoding step, which is usually less flexible than an arbitrary drawing, might lose also some information about the chemical structure, but it will provide a solid starting point for further automated processing. Interesting formats for representing chemical compounds are for example SMILES~\cite{smiles} or MOLfile~\cite{molfile}, which contains all necessary information to build the complete molecule graph structure. Using these formats, it would for example be possible to query documents for specific patterns in chemical compounds. However, such encodings still remains somewhat cumbersome and are not yet systematically available, in particular for unstructured legacy data.

Thousands of scientific publications describe new chemical compounds and investigate their properties. However, the structure of these chemical compounds are usually described in the publication only as an image. This means that today a rich source of data, which would be extremely valuable to develop novel machine learning approaches or simply query documents more accurately, is largely under-exploited.  It is therefore important to convert images of chemical structures into these formats. A few tools for recognizing graph structures from chemical compound images are available, such as OSRA~\cite{osra} and others like ChemReader~\cite{chemreader}, Kekule~\cite{kekule}, CLiDE Pro~\cite{clidepro}, and the work of~\citeauthor{rulebased}. However, we observe that, using these tools, bond multiplicity and stereochemical information are sometimes lost. Those tools are mainly expert systems using different techniques, such as image processing, optical character recognition, hand-coded rules, or sophisticated algorithms. Modifying or further improving these tools requires a lot of effort. A tool based on machine learning, which learns directly from training data, would be most valuable. Such a tool could potentially become more accurate than existing methods and its performance could be improved by increasing the size and the diversity of the data sets, instead of having to modify its code.

Therefore, we propose a new \emph{data-driven} machine learning based tool that can learn from only image data to recognize the chemical structure graph given an image of a chemical structure. The core of the tool is a deep learning model. In the work of \citeauthor{schrodinger}, another deep learning model was also proposed. However, there the output is only a  text-sequence representing the graph. In our approach, we focus on directly predicting the graph structure (\emph{i.e.}, identifying all the nodes and the edges and their labels). The positions of these nodes and edges in the resulting graph would correspond with the positions in the original image of the chemical structure. The resulting graph can be later translated to any format (\emph{e.g.}, SMILES).

In the next sections, we will describe the method and the neural networks used, and also how the different networks interact. Next, we describe the data sets used for training. Then we focus on the performance and scalability of our method, and conclude with possible future work.
\section{Related work and Background}

A deep convolutional neural network \cite{deeplearning} is the type of network most often used for image recognition. These convolutional neural networks can be split into two main types: (1) image segmentation networks and (2) classification networks. We combine both approaches in our graph recognition tool.
\subsection{Image segmentation}
Our work builds upon the recent developments in image segmentation. Different machine learning approaches can be used for the segmentation of images. One well-established approach is U-Net~\cite{unet}. This approach uses a network that combines a contracting path and an expanding path. Several other works were based on the U-net approach, such as~\citeauthor{singing}, where a U-net is used to extract the vocal component from music. Other works expanded this U-net approach, such \citeauthor{3dunet}, which generalizes the U-net approach to 3D images.
Another approach is to make use of dilated convolutions~\cite{dilated} all stacked without loss of resolution. An advantage of dilated convolutions is that the receptive field can grow exponentially by increasing the dilation in the dilated convolutional operator. This would be computationally more interesting than using multiple convolutions or larger kernels. 
\subsection{Image classification}
There has been a trend to create deeper and deeper neural networks to improve performance for the classification of images. However, deeper networks are more difficult to train. To improve the training capabilities of such deep networks, methods such as residual neural networks (Resnet)~\cite{resnet} were developed. It has also been shown that these residual neural network are comparable in performance and behavior with an ensemble of more shallow networks~\cite{resnet_ensenmble}. It is also worth mentioning the work of \citeauthor{resnetunet} where the concept of Resnet is combined with the concept of U-net for the localization of roads on aerial images. 
\subsection{Drug discovery and machine learning}
There are several stages in the process of drug discovery. The stages go from basic research and drug candidate selection to the development phase, clinical trials, and finally production. As development progresses further and sunken costs increase, the cost of failure of a project thus increases. "Fail early" is thus important to contain the costs of drug discovery. Predicting risks of failure later in the discovery process (for example, by predicting toxicity for a compound) without draining the pipeline (enough candidate compounds need to remain available) is essential. Machine learning techniques can be used in all stages of drug discovery. \citeauthor{chen_drug_discovery}  gives a good overview of the recent use of deep learning in drug discovery. We would like to highlight some of these recent applications, which we find interesting in the context of our graph recognition tool. 

In the first place, there is the work from \citeauthor{seq2seqfingerprint} and \citeauthor{smilesautoencode}, where an unsupervised method is used to extract features from SMILES input data. SMILES(Simplified Molecular Identification and Line Entry System)~\cite{smiles} is a text representation of a chemical compound following specific syntactic rules. The unsupervised learning methods in both works are based on the auto-encoder principle. This feature vector can then be used as input to a supervised method to learn to predict molecular properties. 

Another interesting method to predict molecular properties of a chemical compound is to use the neural graph fingerprint presented in \citeauthor{neuralgraphs}. The neural graph fingerprint is a way to represent and encode a chemical compound. Here, a convolutional neural network takes these graphs as input and is trained to predict molecular properties. Similarly, in \citeauthor{kearnes2016molecular}, \citeauthor{convembgraph}, \citeauthor{simm_graph_2019} and \citeauthor{pkcsm}, a machine learning model takes a molecular graph as input.

Large amount of data is needed to use or train the models mentioned above. It is not always easy to find this data. This is where our tool is useful, by extracting graph representations of chemical compounds directly from images. It is also worth mentioning the work presented in \citeauthor{chemception}, where no representation of the chemical compound is needed. In this work, a machine learning model is trained to predict molecular properties directly from images from chemical structures.
\subsection{Stereoisomerism}
Stereochemical information can also be encoded in a 2D representation of a molecule. This stereochemical information is important to differentiate molecules with the same molecular formula but with a different spatial orientation. To encode this central chirality, different type of lines are used to represent bonds in the 2D representation of a molecule: solid lines, wedge-shaped lines or dashed lines~\cite{stereo}. It is important that this information is captured by our graph recognition tool.
\section{Problem Statement}
In this section, we formulate our learning task. The goal of the proposed method is to learn a function that maps an image $\mathbf{x}$ to its graph representation $G$.
\begin{definition}
$\mathbf{x} \in \mathbb{R}^{U \times V}$ represents a single-channel 2D image with dimensions $U \times V$. 
\end{definition}
Generalizations to multiple channels is straightforward, if colored inputs are available.
\begin{definition}
$G = (V,E)$ represents a graph with vertices $V$ and edges $E$.
\end{definition}

For our graph recognition tool to work, we need to learn the following function:
\begin{equation}\label{overall}
    g: \mathcal{X} \rightarrow \mathcal{G}.
\end{equation}
This function will map an 2D input image of a chemical structure to the graph representation of the molecule. To learn this function, we make the following assumption for the training data set:
\begin{assumption}
We assume we have the knowledge about the location of every node in the graph represented by
$\mathbf{L} \in \mathbb{N}^{i \times 2}$, with $i$  number of nodes in the graph for every image in our labeled training data set. We also assume the knowledge of all inter-node connections in the graph (edges) for all labeled images represented by $\mathbf{C} \in \{0,1\}^{i \times i}$.
\end{assumption}
\begin{definition} The training data set $\mathcal{D}$ is then defined as
$\mathcal{D} = \{ (\mathbf{x}_i, \mathbf{L}_i, \mathbf{C}_i) \}_{i=1}^{N_{\mathrm{samples}}}$ .
\end{definition}
\section{Model}
To solve the problem statement defined in the previous section, we build a machine learning model. The model is split up in different learning tasks, which will be defined here. \subsection{First task: segment type segmentation}
The first learning task is to learn to segment a 2D image of a chemical structure in different segments, where each segment can represent the location of a specific atom, charge or bond type in the image. The image was already defined in previous section. However, the segmentation of this image will be defined here.
\begin{definition}
$\mathbf{S}^a \in \mathbb{R}^{U \times V \times n_a}$, $\mathbf{S}^b \in \mathbb{R}^{U \times V \times n_b}$,
$\mathbf{S}^c \in \mathbb{R}^{U \times V \times n_c}$ represent the atom type, bond type and charge segmentation of an image. $U$ and $V$ are the same as in the input image while $n_a$, $n_b$ and $n_c$ respectively are the number of atom types, bond types and charges (including the empty atom, charge and bond type) present in the compound.\end{definition}
To perform image segmentation, we need to learn the following function
\begin{equation}
    s: \mathcal{X} \rightarrow \mathcal{S}^a,{S}^b,{S}^c.
\end{equation}
To learn this function, we need to label the training elements.
\begin{definition} Let $\mathbb{I}_{n} = \{1,...,n\} \subseteq \mathbb{N}$ then we define
$\mathbf{L}^a \in \mathbb{I}_{n_a}^{U \times V}$, $\mathbf{L}^b \in \mathbb{I}_{n_b}^{U \times V}$, $\mathbf{L}^c \in \mathbb{I}_{n_c}^{U \times V}$ which represent the pixelwise true labels. $U$ and $V$ are the same as the input image. $n_a$, $n_b$ and $n_c$ are respectively the number of atom types, bond types and charge (including the empty atom, charge and bond type). The value of every element represent the atom, charge and bond type to which the corresponding pixel belongs.
\end{definition}

Once the true labels for the training data have been defined, we can define the loss function for training. Here, we will use the Cross Entropy Loss, which is defined as
\begin{equation}
    H(y,\hat{y}) = -\sum_{i}^{N}y_i \log \hat{y_i},
\end{equation}
where $y_i$ is the true probability distribution of the true labels, $\hat{y_i}$ is the estimated probability distribution of the labels, and $N$ is the number of different classes. In the case of atom type segmentation $\mathbf{S}^a$, the cross entropy loss is calculated and summed for every pixel prediction (so fixing $u$ and $v$) in the following way, taking into account that $\mathbf{S}^a$ is not a probability distribution:
\begin{equation}
    \mathrm{Loss}_a=
        - \sum_{u=1}^{U}\sum_{v=1}^{V} \log
        \left(
            \frac{\exp(\mathbf{S}^a_{u,v,\mathbf{L}^a_{u,v}})}{
    \sum_{j} \exp(\mathbf{S}^a_{u,v,j})} 
        \right).
\end{equation}
The loss ($\mathrm{Loss}_b$, $\mathrm{Loss}_c$) in the case of bond type segmentation  $\mathbf{S}^b$ and charge segmentation $\mathbf{S}^c$ is calculated similarly. The total loss is the sum of all partial losses:
\begin{equation}
    \mathrm{Loss}_{total}= \mathrm{Loss}_a + \mathrm{Loss}_b + \mathrm{Loss}_c.
\end{equation}

\subsection{Second task: segment type classification}
A second learning task is necessary to build a final graph. This learning task classifies parts of the segmented image into the different possible atom, bond and charge types. One part of the input used in this learning task is defined in the following way:

\begin{definition}
$\mathbf{S}^{a_{cut}} \in \mathbb{R}^{K \times L \times n_a}$, $\mathbf{S}^{b_{cut}} \in \mathbb{R}^{K \times L \times n_b}$, $\mathbf{S}^{c_{cut}} \in \mathbb{R}^{K \times L \times n_c}$  where $K,L = $ 2 times the regular bond length in a 2D image of a chemical structure and $n_a$, $n_b$ and $n_c$ are respectively the number of different atom types, bond types or charges (including the empty types). The tensors $\mathbf{S}^{a_{cut}}$, $\mathbf{S}^{b_{cut}}$ and $\mathbf{S}^{c_{cut}}$  represent a cut-out of the tensors $\mathbf{S}^a$, $\mathbf{S}^b$ or $\mathbf{S}^c$. 
\end{definition}

Another part of the input used in this learning task is defined as:

\begin{definition}
$\mathbf{x}^{a_{cut}} \in \mathbb{R}^{K \times L}$, $\mathbf{x}^{b_{cut}} \in \mathbb{R}^{K \times L}$, $\mathbf{x}^{c_{cut}} \in \mathbb{R}^{K \times L}$ represent the cut-outs of the original 2D image $\mathbf{x} \in \mathbb{R}^{U \times V}$  where $K,L = $ 2 times the regular bond length in a 2D image of a chemical structure. 
\end{definition}

Next, we define the output used in this learning task.

\begin{definition}
$A \in \mathbb{R}^{n_a}$, $B \in \mathbb{R}^{n_b}$, $C \in \mathbb{R}^{n_c}$ are vectors of dimension $n_a$, $n_b$ and $n_c$ where $n_a$, $n_b$ and $n_c$ are respectively the number of different atom types, bond types or charges (including the empty types). The vectors $A$, $B$ and $C$ represent respectively the atom type, bond type and charge predictions. 
\end{definition}
With these definitions, we can now also define the functions to be learned in this task:
\begin{equation}
    c_A : \mathcal{S}^{a_{cut}}, \mathcal{X}^{a_{cut}} \rightarrow \mathcal{A},
\end{equation}
\begin{equation}
    c_B : \mathcal{S}^{b_{cut}}, \mathcal{X}^{b_{cut}} \rightarrow \mathcal{B},
\end{equation}
\begin{equation}
    c_C : \mathcal{S}^{c_{cut}}, \mathcal{X}^{c_{cut}} \rightarrow \mathcal{C}.
\end{equation}
To learn these functions, we need the labels of the training data.
\begin{definition}Let $\mathbb{I}_{n} = \{1,...,n\} \subseteq \mathbb{N}$ then we define
$L_A \in \mathbb{I}_{n_a} $, $L_B \in \mathbb{I}_{n_b} $ and $L_C \in \mathbb{I}_{n_c} $ respectively represent the label of atom type, bond type and charge of each training element. $n_a$, $n_b$ and $n_c$ are respectively the number of atom types, bond types and charge (including the empty atom, charge and bond type).
\end{definition}
Finally, we also define the loss function used in the training phase for learning funtion $c_A$.
\begin{equation}
    \mathrm{Loss}_{c_A} = -\log \left( 
    \frac{\exp(A_{L_{A}})}{\sum_{j} \exp(A_j)} 
    \right)
\end{equation}

The loss functions ($\mathrm{Loss}_{c_B}$, $\mathrm{Loss}_{c_C}$) for the functions $c_B$ and $c_C$ can be defined similarly.
\section{Graph Building Algorithm}
Once we have learned the functions described in the previous section we need an algorithm to combine the outputs of these functions and build up a final graph structure. We propose the algorithm defined in Algorithm~\ref{alg:alg-1}:

\begin{algorithm}[H]
\SetAlgoLined
\KwData{Image tensor $\mathbf{x}$ }
\KwResult{Graph $G$}
 $\mathbf{S}^a$,$\mathbf{S}^b$,$\mathbf{S}^c$  = $s(x)$
 
 $\mathrm{atomCandidates} = \mathrm{generateAtomCandidates}(\mathbf{S}^a)$
 
 $V=$ []
 
 \For{$\mathrm{atomCand}$  \textbf{in} $\mathrm{atomCandidates} $}{
  $\mathbf{S}^{a_{cut}}$,$\mathbf{x}^{a_{cut}}$,$\mathbf{h}^a_L$ = $\mathrm{cutAtomCand}(\mathrm{atomCand}, \mathbf{S}^a, \mathbf{x})$
  
  $\mathbf{S}^{c_{cut}}$,$\mathbf{x}^{c_{cut}}$,$\mathbf{h}^c_L$ = $\mathrm{cutAtomCand}(\mathrm{atomCand}, \mathbf{S}^c, \mathbf{x})$
  
  $A = c_A(\mathbf{S}^{a_{cut}},\mathbf{x}^{a_{cut}}$,$\mathbf{h}^a_L)$
  
  $C = c_C(\mathbf{S}^{c_{cut}},\mathbf{x}^{c_{cut}}$,$\mathbf{h}^c_L)$
  
  \If{$\mathrm{isNotEmptyAtomType}(A)$}{
   $V.\mathrm{appendAtom}(A, C, \mathrm{atomCand})$
   }
 }
 $\mathrm{bondCandidates} = \mathrm{generateBondCandidates}(V)$
 
 $E=$ []
 
 \For{$\mathrm{bondCand}$ \textbf{in} $\mathrm{bondCanidates}$}{
 $\mathbf{S}^{b_{cut}},\mathbf{x}^{b_{cut}}$,$\mathbf{h}^b_L = \mathrm{cutBondCand}(\mathrm{bondCand}, \mathbf{S}^{b}, \mathbf{x}$)
 
 $B = c_B(\mathbf{S}^{b_{cut}},\mathbf{x}^{b_{cut}}$,$\mathbf{h}^b_L)$
 
 \If{$\mathrm{isNotEmptyBondType}(B)$}{
 $E.\mathrm{appendBond}(B, \mathrm{bondCand})$
 
 }
 }
 \caption{Graph building algorithm}
 \label{alg:alg-1}
\end{algorithm}

The proposed algorithm \ref{alg:alg-1} will first apply the segmentation function $s$ to the input image. Next, given the segmentation $\mathbf{S}^a$, candidate locations $\mathrm{atomCandidates}$ will be generated by $\mathrm{generateAtomCandidates}$. Given these candidate locations, the nodes $V$ of the graph can be build in an iterative way. For this purpose, the segmentations $\mathbf{S}^a$ and $\mathbf{S}^c$ can be cut ($\mathrm{cutAtomCand}$) into smaller segments $\mathbf{S}^{a_{cut}}$ and $\mathbf{S}^{c_{cut}}$ thanks to every candidate location $\mathrm{atomCand}$. At the same time the original image $\mathbf{x}$ is also cut ($\mathrm{cutAtomCand}$) into smaller parts $\mathbf{x}^{a_{cut}}$ and $\mathbf{x}^{c_{cut}}$. Extra highlights are also created $\mathbf{h}^a_L$ and $\mathbf{h}^c_L$ which highlight the candidate location to be classified.  Then, the classification functions $c_A$ and $c_C$ are applied to determine what kind of atom and charge is located at the candidate location. If the candidate location is not empty ($\mathrm{isNotEmptyAtomType}$), the location, type $A$ and charge $C$ will be added to the list of nodes $V$. Later, the algorithm will use these nodes $V$ to build the edges $E$ of the graph $G$. For this, it first will need to generate ($\mathrm{generateBondCandidates}$) the candidate bond locations ($\mathrm{bondCandidates}$). Similarly, as for the nodes, the bonds $E$ of the graph can be built in an iterative way. For this purpose, the segmentation $\mathbf{S}^b$ can be cut ($\mathrm{cutBondCand}$) into a smaller segment $\mathbf{S}^{b_{cut}}$ thanks to every candidate bond location $\mathrm{bondCand}$. At the same time the original image $\mathbf{x}$ is also cut ($\mathrm{cutBondCand}$) into a smaller part $\mathbf{x}^{b_{cut}}$. Extra highlights are also created $\mathbf{h}^b_L$ which highlight the bond location to be classified.  Finally, the classification function $c_B$ is applied to determine the type of bond located at the candidate bond location. If the candidate bond location is not empty ($\mathrm{isNotEmptyBondType}$), the location and type $B$ will be added to the list of bonds $E$. 
\section{Deep learning implementation}
 The method we use in the graph recognition tool is a combination of different convolutional neural networks~\cite{deeplearning}. First, we have a semantic segmentation network using the Dense Prediction Convolutional Network~\cite{fcn,dilated} followed by three classification networks. The output of the segmentation network is part of the input of the other classification networks.
 
 \subsection{Semantic segmentation network \texorpdfstring{$s$}{s}}
Before feeding the image to the segmentation network $s$, the image is preprocessed to a binary black and white image. The output of the segmentation network are different channels predicting for every pixels in the image the class the pixels belong to. The possible classes represent the different atom types, bond types and charges. For the implementation of this network, we build on the concept of dilated convolution described in~\citeauthor{dilated}.

\subsubsection{Network architecture}

The network has 8 3x3 convolutional layers from which 6 layers make use of dilation. All convolutional layers are followed by a Rectified Linear Unit (ReLU). The last layer is a linear layer. Padding is used so that the resolution of the channels does not change. The padding and dilation for the different convolutional layers are summarized in Table~\ref{network1}.

\begin{table}
\caption{Summary of the layers of the segmentation network}
\label{network1}       
\begin{tabular}{lllll}
\hline\noalign{\smallskip}
Layer & Kernel & Nonlinearity & Padding & Dilation \\
\noalign{\smallskip}\hline\noalign{\smallskip}
     conv1 & 3x3 & ReLU & 1 & no dilation \\
     conv2 & 3x3 & ReLU & 2 & 2 \\
     conv3 & 3x3 & ReLU & 4 & 4 \\
     conv4 & 3x3 & ReLU & 8 & 8 \\
     conv5 & 3x3 & ReLU & 8 & 8 \\
     conv6 & 3x3 & ReLU & 4 & 4 \\
     conv7 & 3x3 & ReLU & 2 & 2 \\
     conv8 & 3x3 & ReLU & 1 & no dilation \\
     last & 1x1 & none & no padding & no dilation \\
\noalign{\smallskip}\hline
\end{tabular}
\end{table}

\subsection{Classification networks}
For the atom location, the bond prediction and the charge prediction we use three separate classification networks. All three networks use part of the output of the segmentation networks in their input.

\subsubsection{Atom location prediction \texorpdfstring{$c_A$}{cA}}
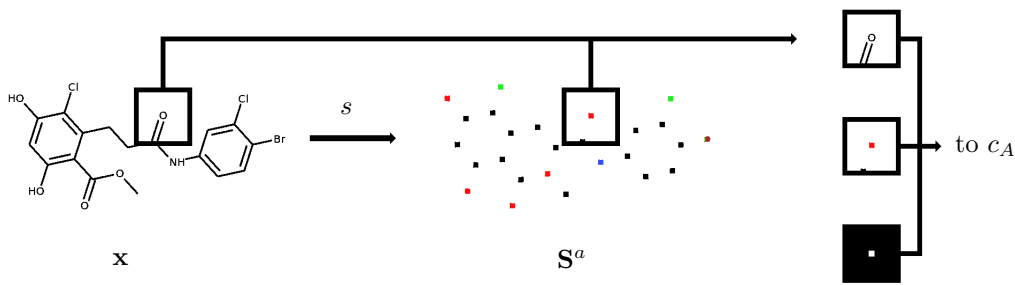
\begin{figure}
    \centering
    \begin{tikzpicture}
    \draw (0, 0) node[inner sep=0]{\includegraphics[ width=13.5cm]{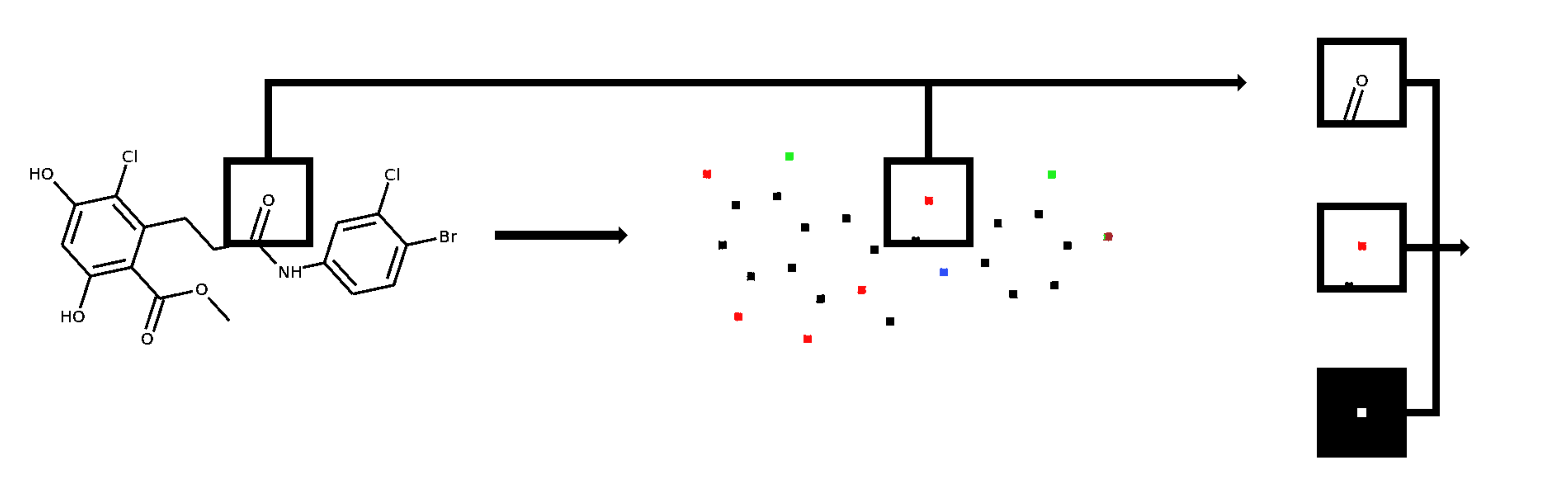}};
    \draw (1, -1.5) node {$\mathbf{S}^a$};
    \draw (-5,-1.5) node {$\mathbf{x}$};
    \draw (-2,0.5) node {$s$};
    \draw (6.5,0) node {to $c_A$};
    \end{tikzpicture}
    \caption{To build the input for the atom classification network ($c_A$), the output of the segmentation network $\mathbf{S}^a$ is cut ($\mathrm{cutAtomCand}$) to feed it in to the atom classification network. This cut-out is shown in the middle ($\mathbf{S}^{a_{cut}}$). To this, we also add part of the original image ($\mathbf{x}^{a_{cut}}$) together with highlighting the candidate location ($\mathbf{h}^a_L$) of the atom type to be classified. The complete input for $c_A$ is shown on the right. }
    \label{fig:input_atom_pred}
\end{figure}

For the atom location prediction, part of the output the segmentation network ($\mathbf{S}^a$) is used. This output contains the segmentation of only the atoms in the image. Next, the original binary image ($x$) is also used for the input. Finally, candidate locations are created ($\mathrm{generateAtomCandidates}$) to spot the part of the image we want to classify. This is also formatted as input and fed into the network. Depending on the candidate location, the inputs can be reduced ($\mathrm{cutAtomCand}$) so that only the immediate region of the candidate location is included ($\mathbf{S}^{a_{cut}}$). This will reduce the computational cost and speed up the learning of the network. This process is illustrated in Figure~\ref{fig:input_atom_pred}.

The output of this network is a vector ($A$) with every element representing the prediction of the network. The size of this vector is the number of different atom classes, plus one to represent the empty class (no atom). For every image segmentation, this network has to run several times to classify all candidate locations to get all atom predictions in the original image.

\subsubsection{Bond prediction \texorpdfstring{$c_B$}{cB}}
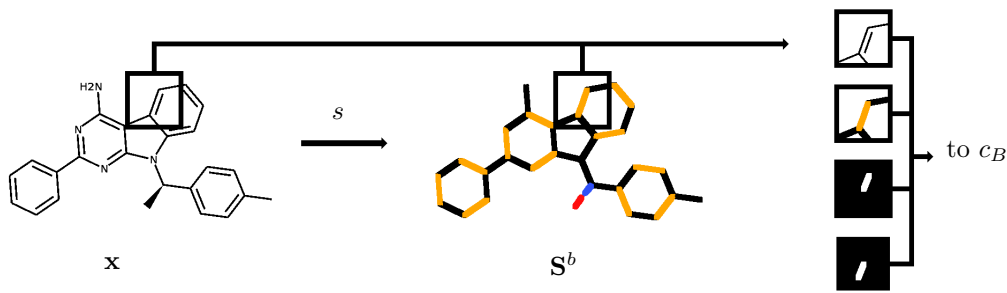
\begin{figure}
    \centering
    \begin{tikzpicture}
    \draw (0, 0) node[inner sep=0]{\includegraphics[width=13.5cm]{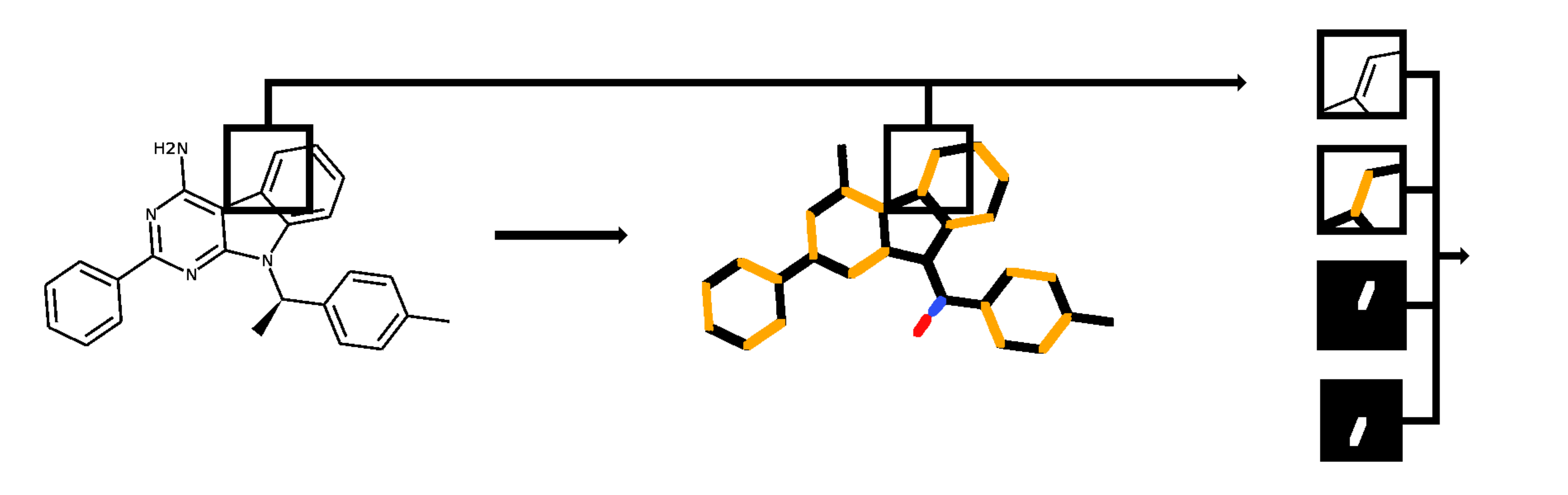}};
    \draw (1, -1.5) node {$\mathbf{S}^b$};
    \draw (-5,-1.5) node {$\mathbf{x}$};
    \draw (-2,0.5) node {$s$};
    \draw (6.5,0) node {to $c_B$};
    \end{tikzpicture}
    \caption{To build the input for the bond classification network ($c_B$), part of the segmented image ($\mathbf{S}^b$) is cut out ($\mathrm{cutBondCand}$). This cut-out ($\mathbf{S}^{b_{cut}}$) is shown in the middle of the figure. To this, we also add part of the original binary input image ($\mathbf{x}^{b_{cut}}$) and the candidate bond location ($\mathbf{h}^b_L$) encoded in two parts. The complete input for $c_B$ is shown on the right.}
    \label{fig:input_bond_pred}
\end{figure}

For the bond prediction network, we apply a similar strategy as illustrated in Figure \ref{fig:input_bond_pred}. This time another part of the output the segmentation network is used ($\mathbf{S}^b$). This output contains the segmentation of only the bonds in the image. Every type of bond is represented in the segmentation as a rectangle. For stereo bonds, we use two rectangles to encode the direction of the bond. Next, as in the atom prediction network, the original binary image ($x$) is used also for the input of the bond prediction network. Finally, for the bond prediction, we also need to encode candidate locations for the bond predictions. This time, as opposed to the atom prediction network, we will use two parts. One rectangle represents the first part of the bond connected with the first atom and another rectangle represents the second part connected with the second atom. The rectangles meet in the middle. By using two rectangles we can encode the direction of the bond which is necessary to predict the stereoisomeric bond direction. These candidate locations are generated ($\mathrm{generateBondCandidates}$) from the predictions from the atom location network. Moreover, depending on these locations, we can cut out ($\mathrm{cutBondCand}$) the inputs again so that only the immediate region of these candidate pairs is fed into the network. 

Like with the atom prediction network, the output is again a vector ($B$) with every element representing the prediction of the network. This time the vector size is the number of different bond classes, plus one to represent the empty class (no bond). 

\subsubsection{Charge prediction \texorpdfstring{$c_C$}{cC}}

\begin{figure}
    \centering
    \begin{tikzpicture}
    \draw (0, 0) node[inner sep=0]{\includegraphics[width=13.5cm]{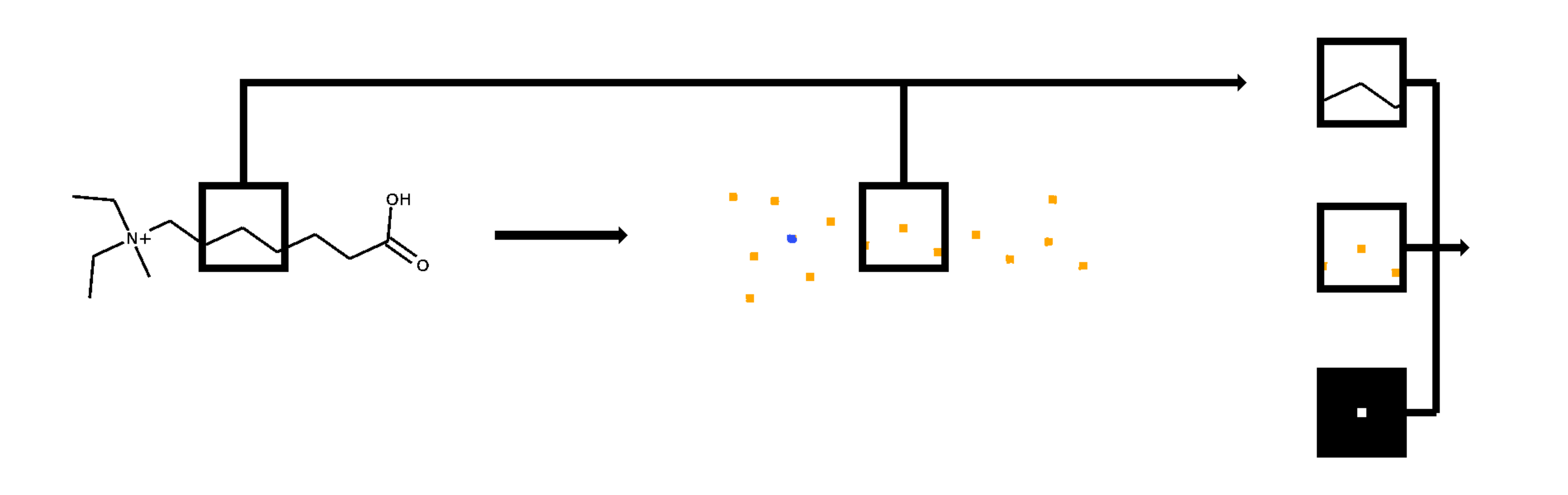}};
    \draw (1, -1.5) node {$\mathbf{S}^c$};
    \draw (-5,-1.5) node {$\mathbf{x}$};
    \draw (-2,0.5) node {$s$};
    \draw (6.5,0) node {to $c_C$};
    \end{tikzpicture}
    \caption{To build the input for the charge classification network ($c_C$), the output of the segmentation network $\mathbf{S}^c$ is cut ($\mathrm{cutAtomCand}$) to feed it in to the charge classification network. This cut-out is shown in the middle ($\mathbf{S}^{c_{cut}}$). To this, we also add part of the original image ($\mathbf{x}^{c_{cut}}$) together with highlighting the candidate location ($\mathbf{h}^c_L$) of the charge type to be classified. The complete input for $c_C$ is shown on the right. }
    \label{fig:input_charge_pred}
\end{figure}
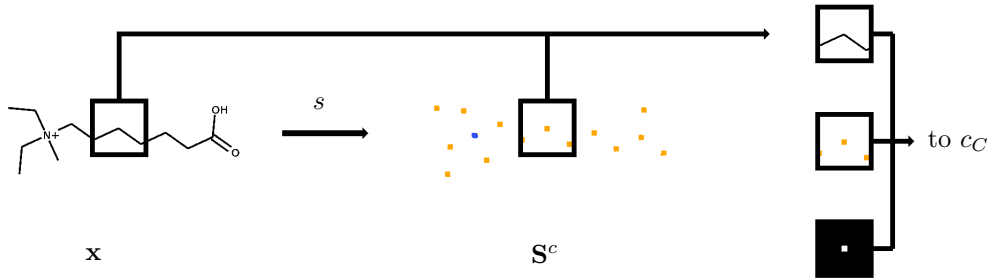

As with the bond prediction and the atom prediction network a similar strategy is used illustrated in Figure \ref{fig:input_charge_pred}. Again a part of the output the segmentation network is used ($\mathbf{S}^c$). This output contains the segmentation of only the charges in the image. Every charge is represented by a rectangle located on the location of the atom it applies to. Depending on the candidate location generated by $\mathrm{generateAtomCandidates}$, the inputs can be again reduced ($\mathrm{cutAtomCand}$) so that only the immediate region of the candidate location is included ($\mathbf{S}^{c_{cut}}$). Again the original image ($x$) is fed together with the candidate location as input to the charge prediction network.

Like with the atom and bond prediction network, the output is again a vector ($C$) with every element representing the prediction of the network. This time the vector size is the number of different charge classes, plus one to represent the empty class (no charge). 

Once we have the bond predictions together with the atom and charge predictions, we can build the graph structure of the complete molecule.

\subsubsection{Network architecture}
The three classification networks have similar layer structures.
There are 5 convolutional layers where 3 of them are dilated and 1 of them (the first one) is actually a depthwise separable convolution \cite{Chollet_2017_CVPR}. After the convolutional layers, there is always a ReLU layer. The last layer is a linear layer and the layer before that is a max pool layer. All layers are summarized in Table~\ref{table:network2}.

\begin{table}
\caption{Different layers in the classification network}
\label{table:network2}      
\begin{tabular}{lllll}
\hline\noalign{\smallskip}
Layer & Kernel & Nonlinearity & Padding & Dilation \\
\noalign{\smallskip}\hline\noalign{\smallskip}
     depthconv1 & 3x3 & ReLU & 1 & no dilation \\
     conv2 & 3x3 & ReLU & 2 & 2 \\
     conv3 & 3x3 & ReLU & 4 & 4 \\
     conv4 & 3x3 & ReLU & 8 & 8 \\
     conv5 & 3x3 & ReLU & 1 & no dilation \\
     maxpool & 124x124 & None & no padding & no dilation \\
     last & 1x1 & None & no padding & no dilation \\
\noalign{\smallskip}\hline
\end{tabular}
\end{table}
\section{Data Sets}

To build our data sets for the segmentation network and the classification networks we download and split different chemical structures in SMILES format from the ChEMBL \cite{chembl_2017} database. The around 1.9 million chemical structures are splitted in 4 parts: 

\begin{itemize}
\item a training pool for the segmentation network of 1.5 million chemical structures, 
\item a pool of 300K chemical structures used for the validation of the segmentation network and training of classification networks, 
\item a pool of around 35K chemical structures for the validation of the classification networks and 
\item another pool of 35K chemical structures for testing the overall performance.
\end{itemize}

From these pools we can sample the actual data sets for our different networks. By sampling we can control the relative frequency of different atom types and bond types in the actual data sets. This is important for the performance of our networks, due to data imbalance, on the different atom types and bond types as we will see in the next section.

\subsection{Segmentation Data Set}
For the training of the segmentation network ($s$), we need 2D images of chemical structures together with pixelwise labeled target values ($\mathbf{L}^a$, $\mathbf{L}^b$ and $\mathbf{L}^c$). This type of data set is not available as far as we know, so we need to construct this data set ourselves. Labeling thousands of 2D images of chemical structures pixelwise by hand is moreover not feasible. We thus construct an automatic procedure to generate this data set. For the training data set we sample around 114K chemical compounds in SMILES format from the ChEMBL training pool in a way that every atom type is present in at least 1000 chemical compounds. Using RDkit~\cite{rdkit} in Python, we create the images starting from the SMILES. Furthermore, to create the labelling, we make some modifications in the code of RDkit at the drawing time of the image, so that it additionally produces the necessary labeling information needed to create our data set. For the validation dataset the same procedure can be used.

\subsection{Atom prediction data set}
Once the segmentation network is trained and validated we can sample from the ChEMBL data a new data set to feed into the segmentation network. The output of these runs are saved to create the input data set for the next classification networks. As already explained in the previous section, the atom prediction network ($c_A$) additionally expects as input the candidate locations to classify. For the training and validation data sets of the classification network $c_A$, we generate candidate locations based on the true atom location values, but also add locations where no atom is located for the prediction of the empty class. For these locations, we take the middle point of every bond in the data set. As we know that no atom is located in the middle of a bond, these locations can be used for the empty values in the data sets.

\subsection{Bond prediction data set}
For the bond prediction network ($c_B$), we apply a similar technique. In addition to the inputs from previous segmentation network, the bond prediction network expects the candidate bond locations. For the training and validation data of $c_B$, we generate these candidate locations by going over all possible combinations of pairs of atoms in a molecule within the range of less than two times the  bond length. Sometimes there is a bond between a generated pair of atoms and then the data set item is labeled with the type of bond. If there is no bond between a pair of atoms, the item is labeled as empty.

\subsection{Charge prediction data set}

For the charge classification network the same data sets as the atom prediction data sets can be used except for the labels. Instead of the atom types the labels would now be the charge (including empty charge) of the atom candidate.
\section{Experiments and results}

For validation we sample new data sets from the validation ChEMBL pools for the different networks. For the segmentation network we sample around 12K chemical structures. For the validation of the classification networks less chemical structures are needed, so we only sample around 450 chemical structures. Starting from these 450 chemical structures we generate atom candidates and bond candidates. This results in a data set of around 27K of atom candidates for atom type and charge classification networks and a data set of around 55K of bond candidate locations for the bond classification network. With these validation data sets we measure the performance on the different networks.

\subsection{Performance of segmentation network}
\begin{figure}
\begin{subfigure}{.5\linewidth}
\centering
\includegraphics[scale=0.45]{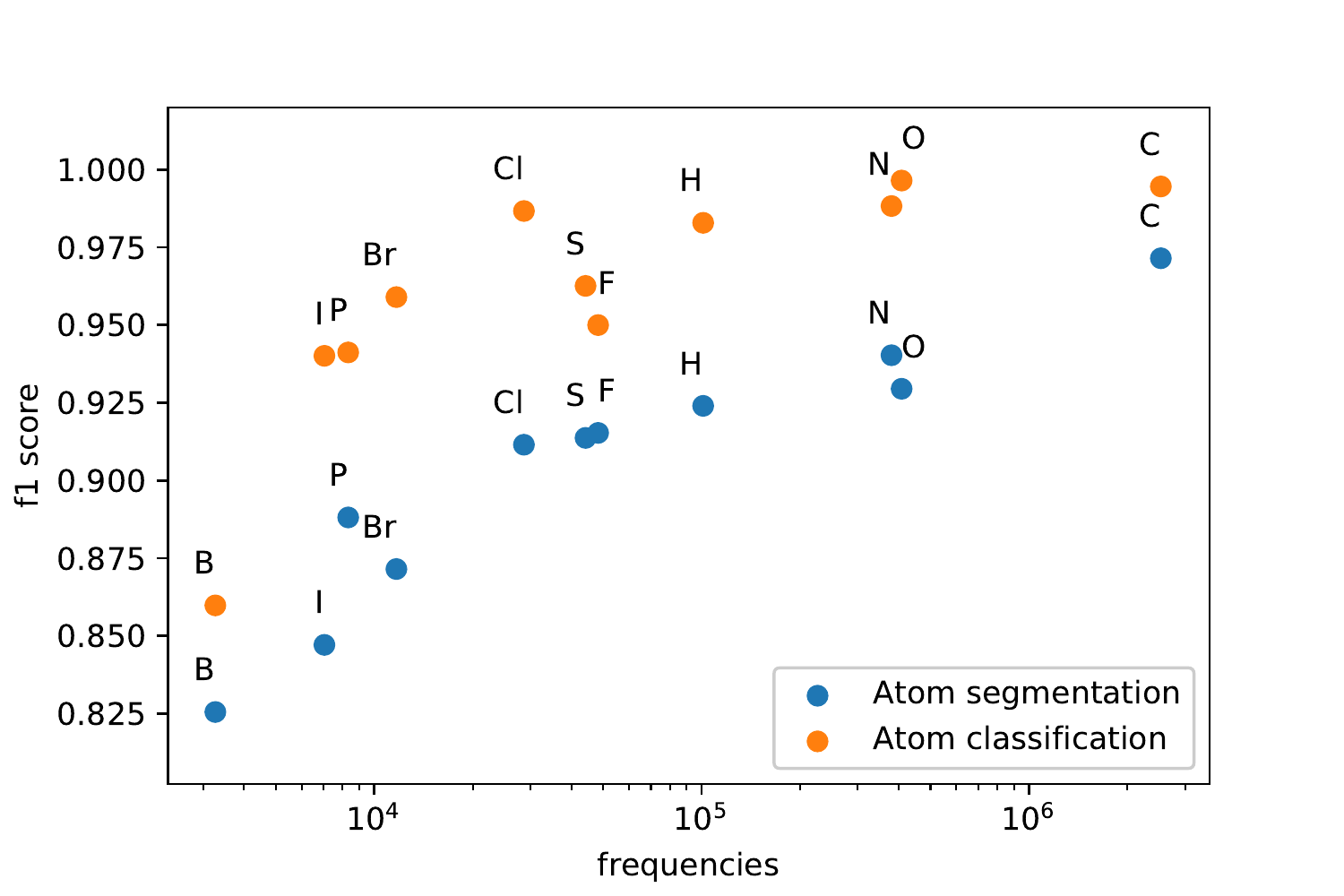}
\caption{Atom prediction performance ($s_A$ and $c_A$)}
\label{fig:sub1}
\end{subfigure}%
\begin{subfigure}{.5\linewidth}
\centering
\includegraphics[scale=0.45]{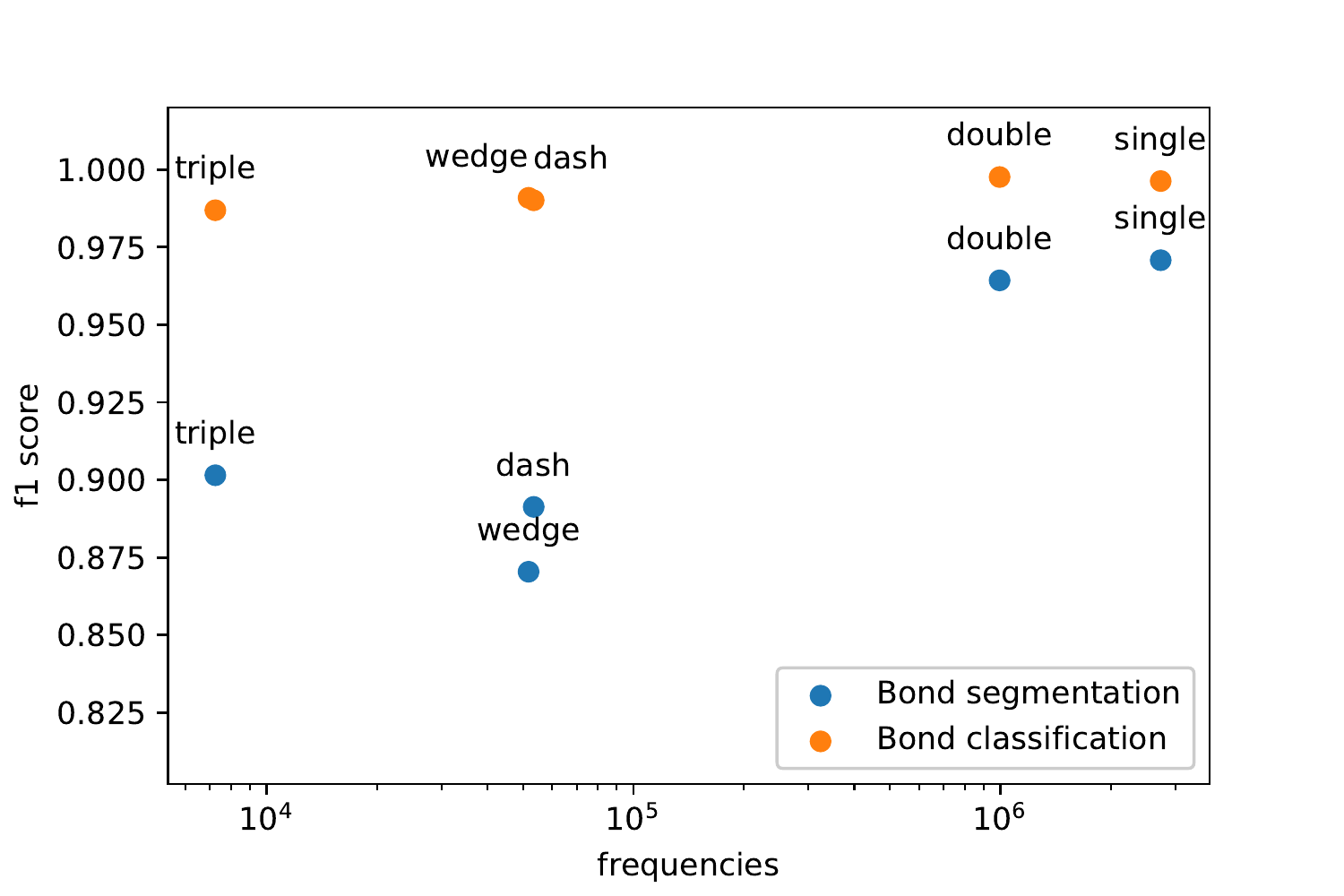}
\caption{Bond prediction performance ($s_B$ and $c_B$)}
\label{fig:sub2}
\end{subfigure}\\[1ex]
\begin{subfigure}{\linewidth}
\centering
\includegraphics[scale=0.45]{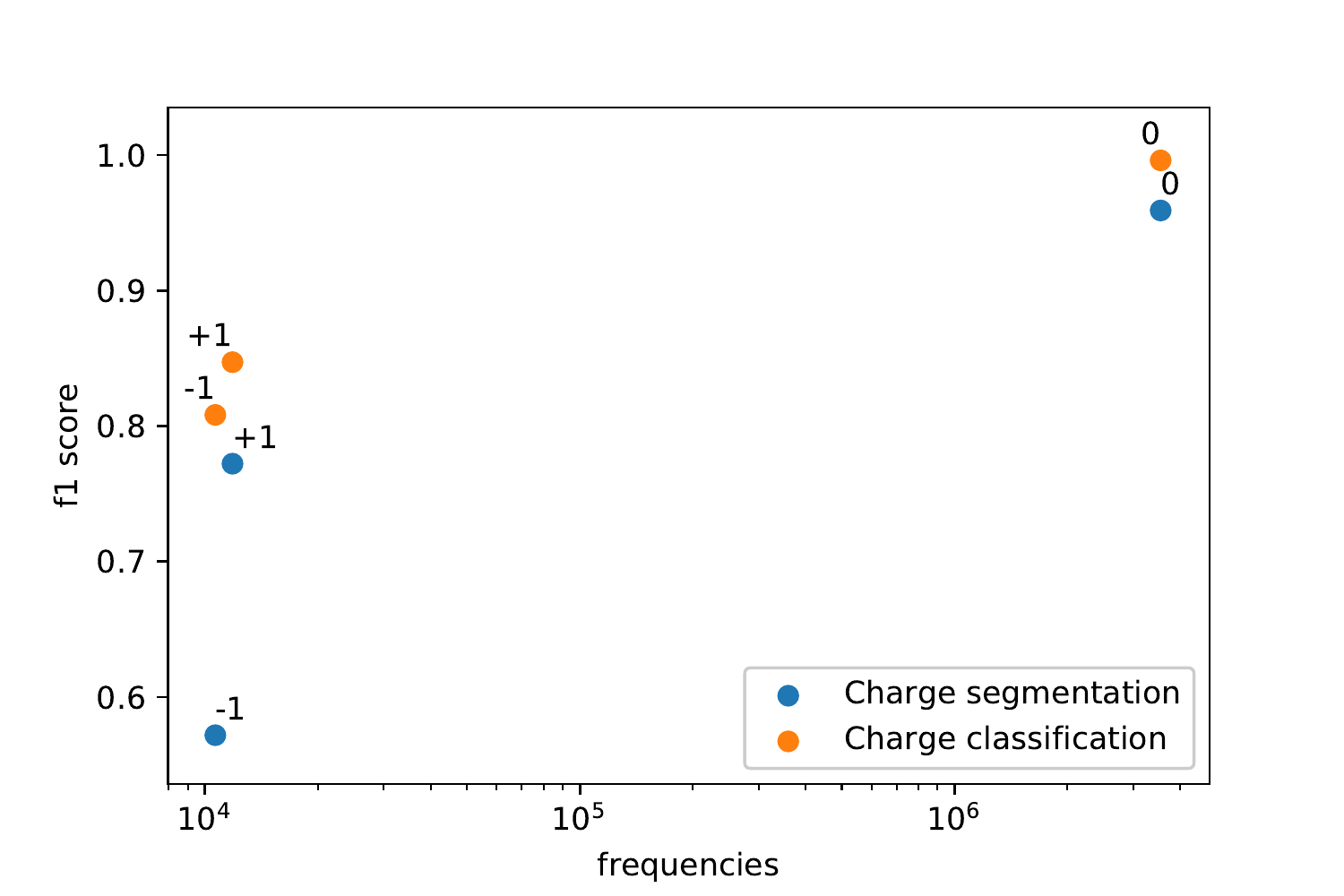}
\caption{Charge prediction performance ($s_C$ and $c_C$)}
\label{fig:sub3}
\end{subfigure}
\caption{F1 score for segmentation and classification networks. There is clearly a correlation between the performance of the networks on the different prediction types and the frequency of the specific type in the training data set. The classification networks perform significantly better than the segmentation networks. }
\label{fig:network_results}
\end{figure}

For the segmentation network ($s$) we measure the F1 score \cite{noauthor_f1_2020} for all the pixel predictions for the different atom, bond and charge types. The F1 score takes into account both precision and recall equally. If we compare the F1 score with the frequency of the different atom, bond and charge types in the training data set we clearly see a correlation. The results are summarized in Figure \ref{fig:network_results} . 

\subsection{Performance of classification networks}
For the classification networks we again use the F1 score to measure the performance for the atom, bond and charge type classifications. Again we see a correlation between the F1 score and the frequency of the different types in the training data set. We can also empirically see that the F1 score for the classification networks is significantly higher than for the segmentation networks. So the classification networks can do a good job even when the segmentation is not perfect. The performance of these classification networks have to be very good as for  every  graph  prediction  tens  of  bond  and  atom  classifications  have  to  made  and  this  would  otherwise  degrade  the  overall accuracy rapidly. The results are summarized in Figure \ref{fig:network_results} .

\subsection{Overall graph accuracy}

\begin{figure}
\begin{subfigure}{\linewidth}
\includegraphics[scale=0.4]{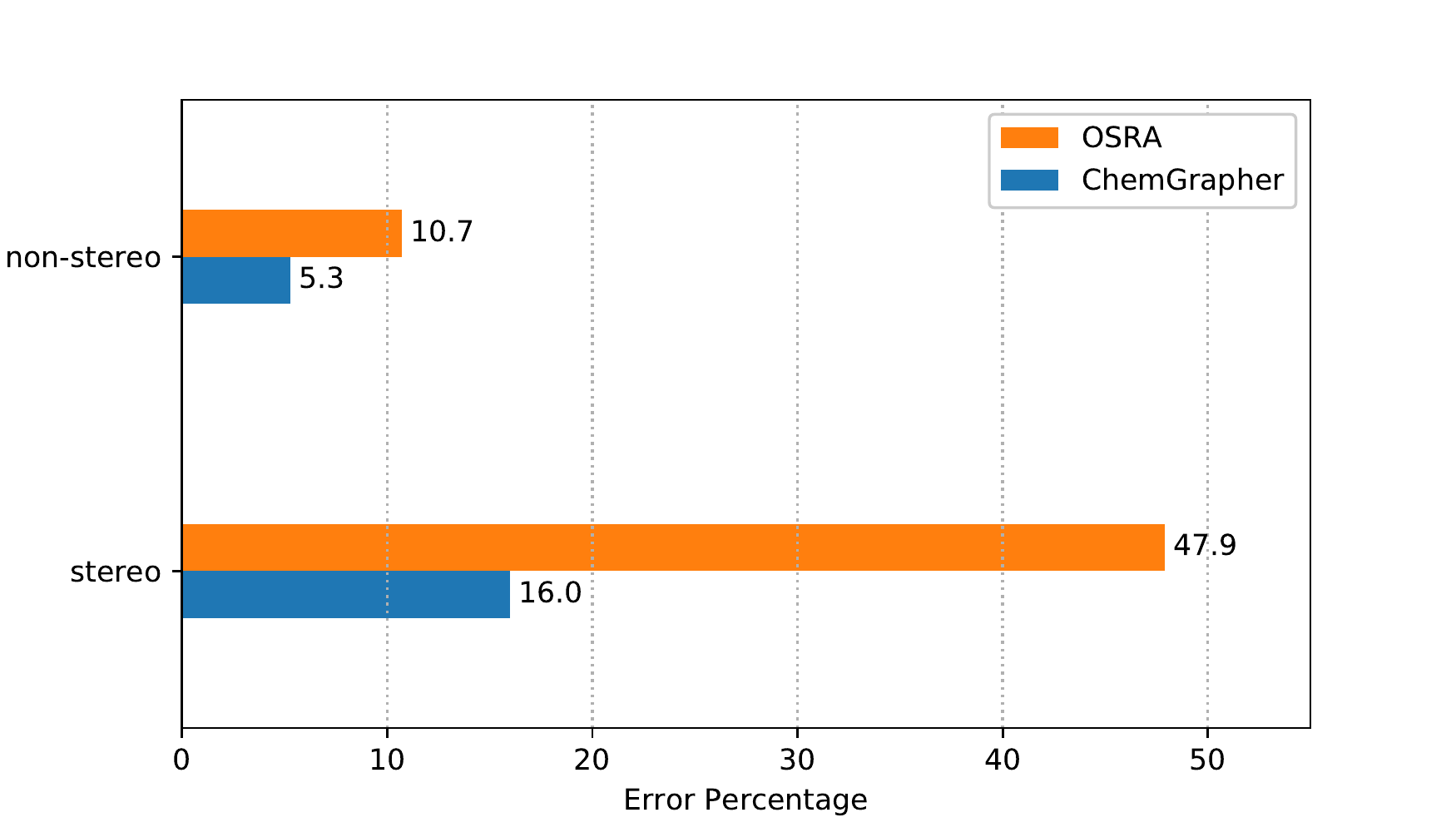}
\includegraphics[scale=0.4]{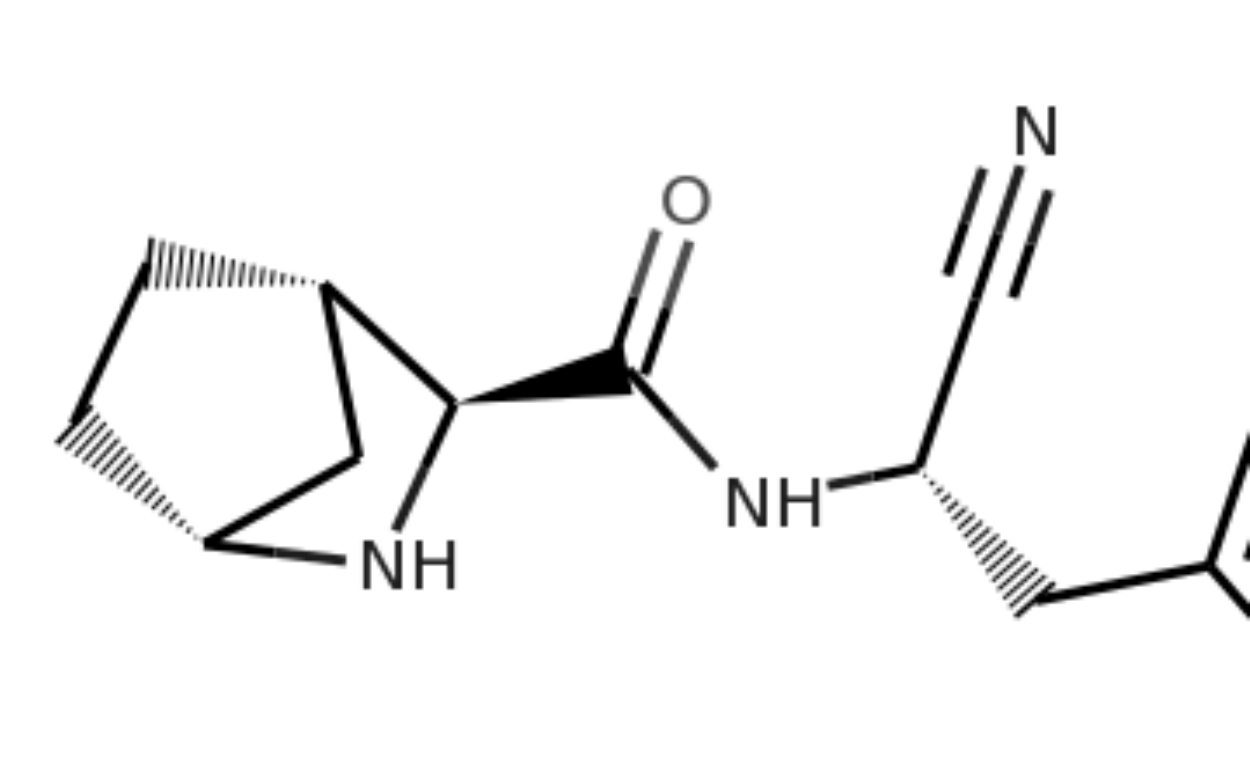}
\caption{Error rate for style 1}
\label{fig:style_sub1}
\end{subfigure}

\begin{subfigure}{\linewidth}
\includegraphics[scale=0.4]{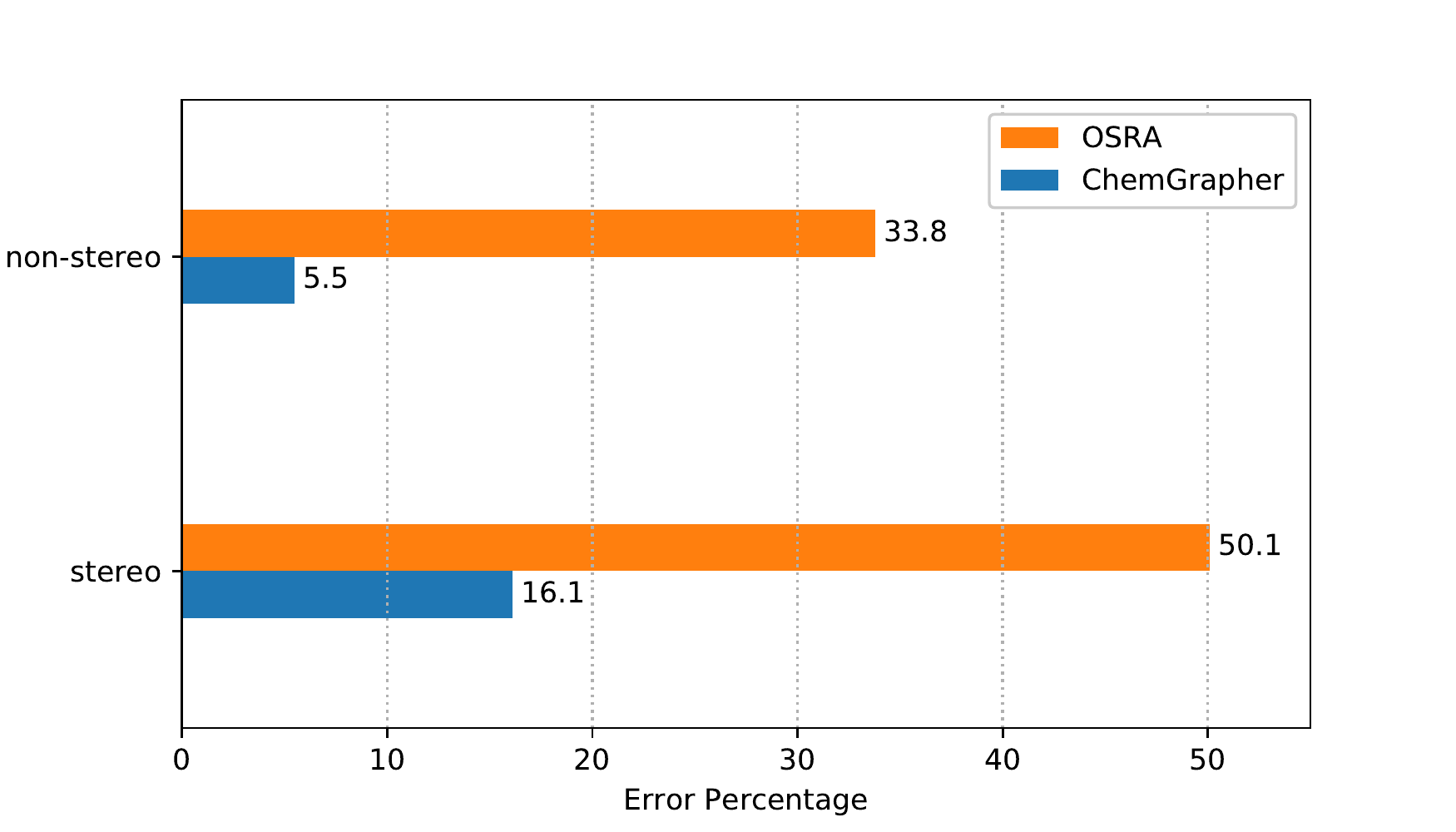}
\includegraphics[scale=0.4]{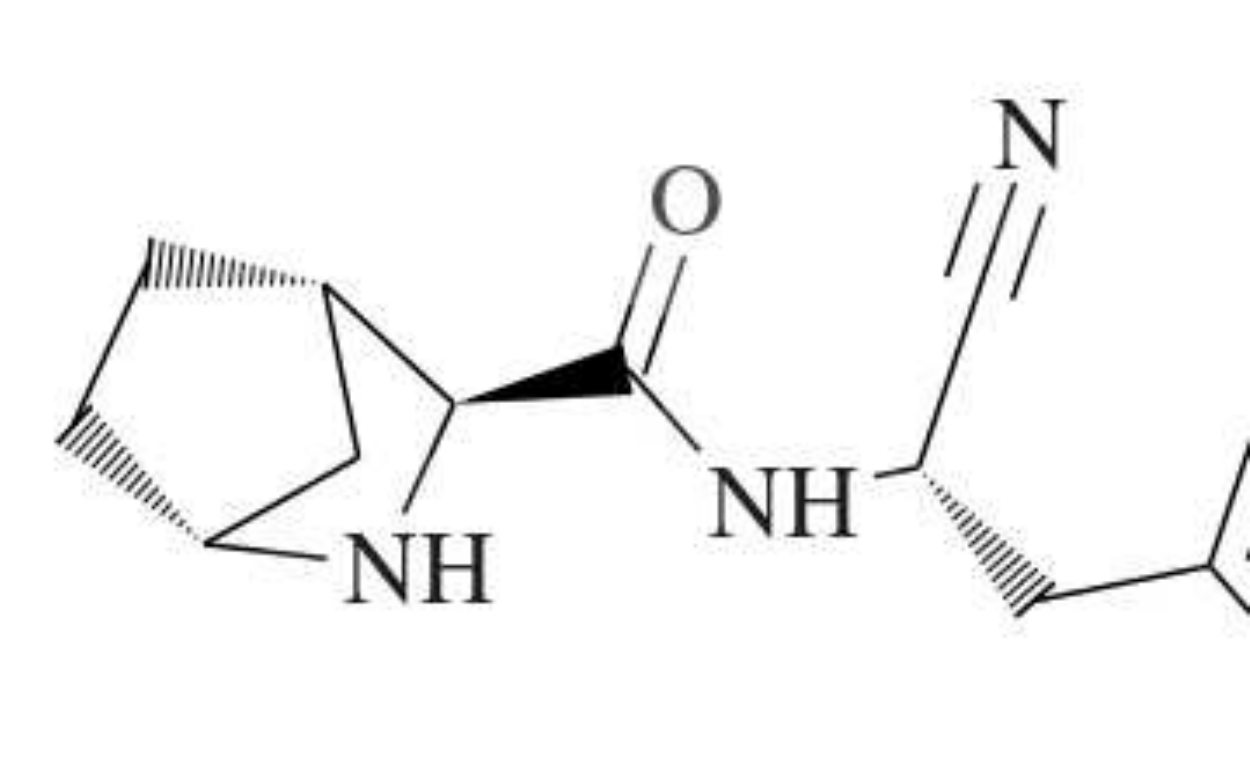}
\caption{Error rate for style 2}
\label{fig:style_sub2}
\end{subfigure}

\begin{subfigure}{\linewidth}
\includegraphics[scale=0.4]{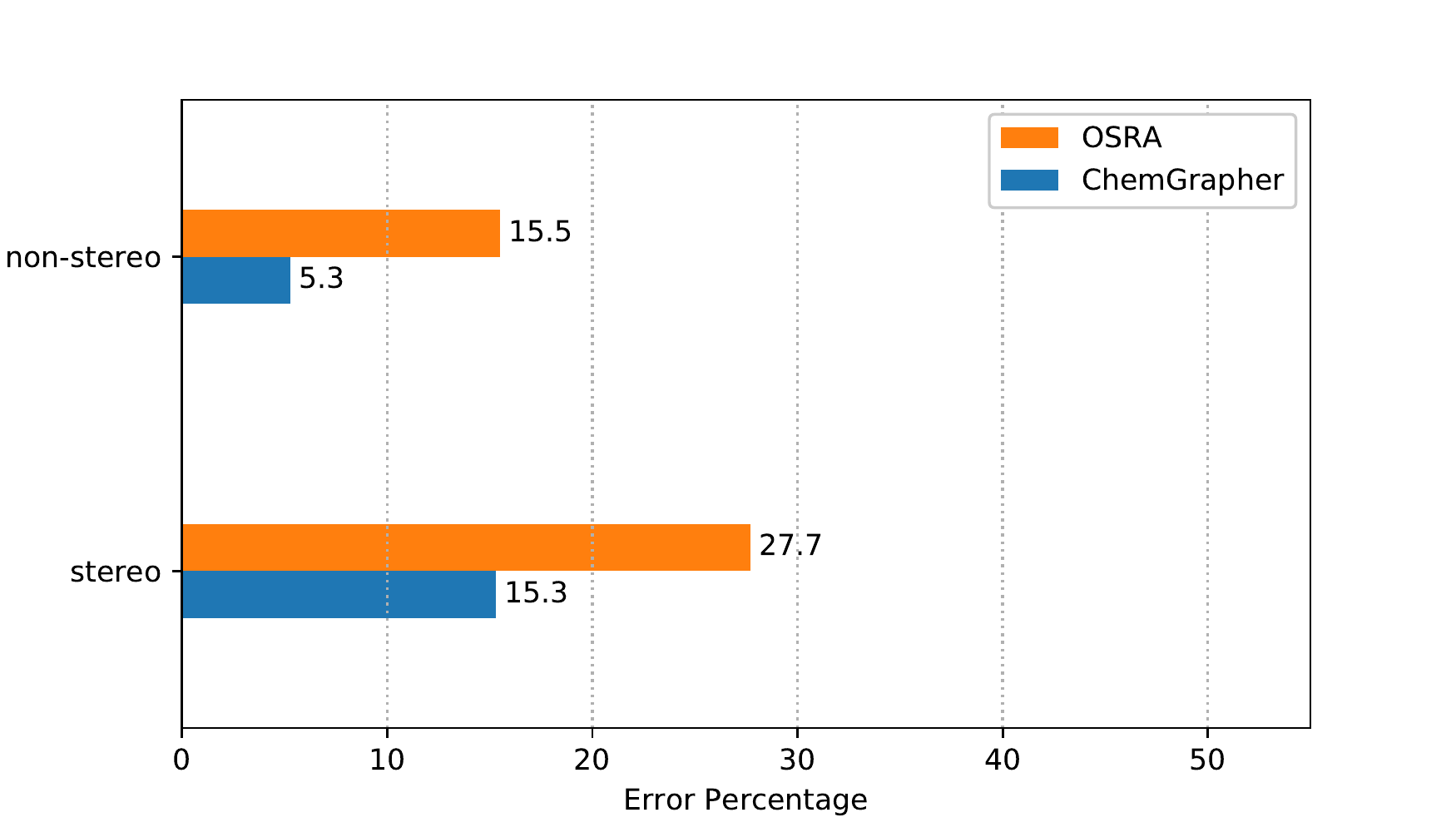}
\includegraphics[scale=0.4]{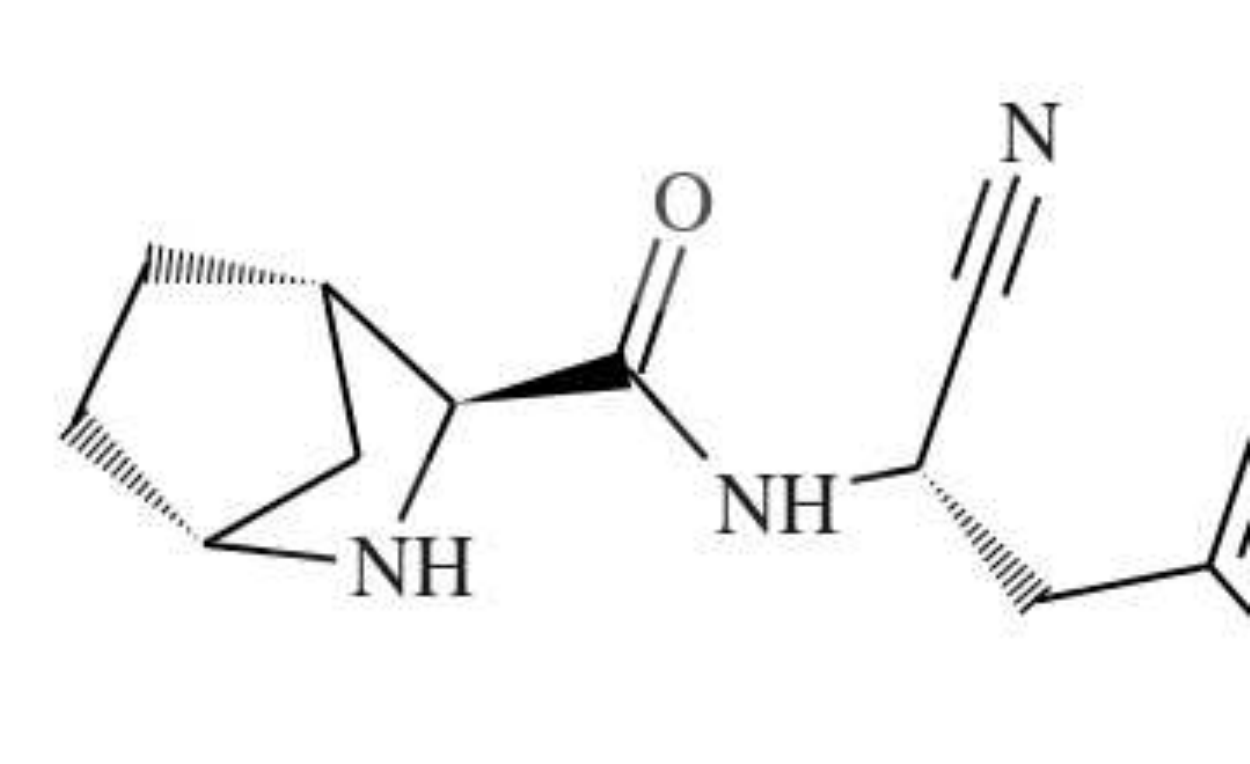}
\caption{Error rate for style 3}
\label{fig:style_sub3}
\end{subfigure}

\caption{The graph accuracy of our tool compared with OSRA measured on images generated in different styles. For each style 2 experiments are performed: once with images without stereo-chemical information and once with images with stereo chemical information. On the left we observe for each style the results on the error rate and on the right we observe for each style an example image in that specific style. For all styles we measure a lower error rate for our tool ChemGrapher compared to OSRA. }
\label{fig:overall_results}
\end{figure}

Now that we know the performance of the different parts, we can combine those building blocks and measure the overall accuracy  of  the  resulting  graph  predictions.  As  already  mentioned  in  a  previous  section,  the  segmentation  network  and classification  networks  should  be  used  as  presented  in algorithm \ref{alg:alg-1}  in order to build the resulting graph. Images in 3 different styles are generated and for every style we generate 2 sets where 1 set only has images without stereo chemical information encoded in the compounds while the other set has images where all compounds have stereo chemical information encoded. This results in 6 sets of each 1000 images to measure the performance on our tool ChemGrapher. If we count at least one mistake in the resulting graph we count the graph prediction as incorrect. The same sets we also use to measure the performance on OSRA to compare. The results are summarized in Figure \ref{fig:overall_results} . On all sets we observe a a higher accuracy on our tool ChemGrapher compared to OSRA.

\subsection{Case Study: Performance on Journal Article Images}

\begin{figure}
\includegraphics[scale=0.4]{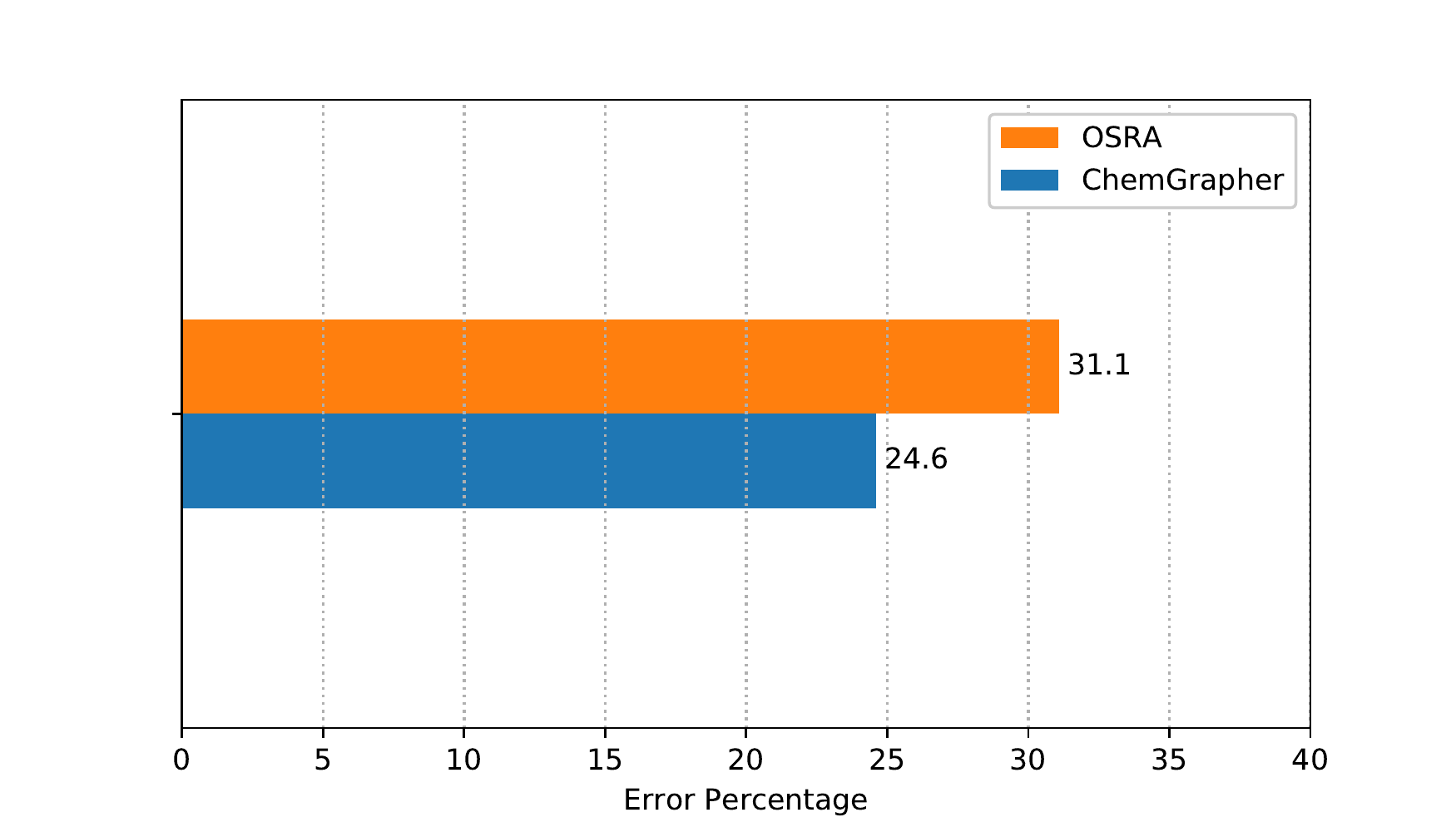}
\caption{Error rate of our tool ChemGrapher on test set of journal article images compared with OSRA. From the errors we learn there is still room for improvement in future work.}
\label{fig:real_images}
\end{figure}

The idea of ChemGrapher is to use it on images in journal articles therefore we would also like to know how well this tool performs on such images. As this kind of data set is not available we decided to build one manually. So we cut out images from journal articles about chemical compounds, preprocess them to the correct input format and feed them to our tool. We evaluate the resulting graph manually on correctness to measure the accuracy. If we count at least one mistake in the resulting graph we categorize the prediction as incorrect. The same procedure was executed for OSRA for comparison. The results of this experiment are summarized in Figure~\ref{fig:real_images}. Thus, out of total of 61 images we tried on ChemGrapher, 46 were correctly predicted while OSRA predicted 42 images correctly. However, we can also observe that ChemGrapher clearly has better performance on images of compounds with only carbon atoms compared to OSRA. For these compounds typically no letters appear in the image. Another observation we make is that ChemGrapher still has some issues related when thick lines are used to depict the bonds. We set this as a target for our future work. 

\section{Future Work}
To train the segmentation network, we need a pixelwise labeled data set. However, this kind of data set is not always available. We thus created this data set with RDkit. However the consequence is that the format of the input image is somewhat biased. We have seen in the case study that ChemGrapher performs reasonably although not equally well on real images. To handle other kind of image formats, it might be difficult to find a pixelwise labeled data set to retrain our networks. Therefore, future work could focus on building a method that can learn from data that is not labeled pixelwise. The data would only offer a way to verify if the resulting graph is correct or not. We could consider this as an instant of weakly supervised learning. 
\section{Conclusion}
We presented a method to recognize the graph structure of molecules from 2D images of chemical structures using deep learning. This method learns a model directly from data. We have seen that careful data preparation is crucial. Care should be taken to have a balanced data set for the different classes of atoms and bonds. However, even with an imperfectly balanced data set, our deep learning methods give very good results. One thing that is important for our method to work is the fact that the classification networks need to have an almost perfect accuracy. While the segmentation network can tolerate some errors, for the classification networks every drop in accuracy can have dramatic results on the overall accuracy. The performance is also clearly better than the well known tool OSRA~\cite{osra} and also provides us more detailed information about the resulting graph.
For our deep learning method to learn accurately, we also had to implement an automatic procedure to pixelwise label 2D images of chemical structures. We described how we modified the code of RDkit for this purpose. In fact, this pixelwise labelling of images for the segmentation is actually key to linking the atoms and bonds in the resulting graph back to the source image. This makes this deep learning method non-black box or interpretable.
In the context of drug discovery, such tools are important. In fact, in general we see that machine learning is gaining importance in this area and that it contributes to improving the quality of the drug discovery process.

\subsubsection*{Acknowledgments}
MO, AA, YM and JS are funded by (1) Research Council KU Leuven: C14/18/092 SymBioSys3; CELSA-HIDUCTION, (2) Innovative Medicines Initiative: MELLODY, (3) Flemish Government (ELIXIR Belgium, IWT: PhD grants, FWO 06260) and (4) Impulsfonds AI: VR 2019 2203 DOC.0318/1QUATER Kenniscentrum Data en Maatschappij. Computational resources and services used in this work were provided by the VSC (Flemish Supercomputer Center), funded by the Research Foundation - Flanders (FWO) and the Flemish Government – department EWI. We also gratefully acknowledge the support of NVIDIA Corporation with the donation of the Titan Xp GPU used for this research.



\end{document}